\documentclass[lettersize,journal]{IEEEtran}
\usepackage{amsmath,amsfonts}
\usepackage{algorithmic}
\usepackage{algorithm}
\usepackage{array}
\usepackage[caption=false,font=normalsize,labelfont=sf,textfont=sf]{subfig}
\usepackage{textcomp}
\usepackage{stfloats}
\usepackage{url}
\usepackage{verbatim}
\usepackage{graphicx}
\usepackage{cite}
\usepackage{subcaption}
\usepackage{xcolor}
\usepackage{hyperref}
\usepackage{wrapfig}
\usepackage{romannum}
\usepackage{booktabs}
\usepackage{svg}
\hypersetup{%
  colorlinks=true,%
  linkcolor={blue},
  citecolor={blue},
  urlcolor={blue!80!black},
  bookmarksnumbered=true,%
  bookmarksopen=true}

\renewcommand{\IEEEauthorrefmark}[1]{\textsuperscript{#1}}

\begin{document}

\title{A Master-Slave Robot Manipulator for Needle-Based Teleoperation in MRI Chamber}

\author{\IEEEauthorblockN{Omar Curiel\IEEEauthorrefmark{1}, Jing-Yuan Huang\IEEEauthorrefmark{1}, Po-Chih Chen\IEEEauthorrefmark{1}, Ji Ma\IEEEauthorrefmark{2}, Qing Dai\IEEEauthorrefmark{4}, Wenqi Zhou\IEEEauthorrefmark{4}, David Lu\IEEEauthorrefmark{3}, Holden H. Wu\IEEEauthorrefmark{4}, Tsu-Chin Tsao\IEEEauthorrefmark{1}}
\thanks{
\IEEEauthorblockA{\IEEEauthorrefmark{1} Department of Mechanical and Aerospace Engineering, Samueli School of Engineeering, University of California, Los Angeles, CA 900095 USA (email: omarzcuriel14@g.ucla.edu;jykevinhuang@g.ucla.edu;pcchen0106@g.ucla.edu;\\ttsao@ucla.edu)}
\IEEEauthorblockA{\IEEEauthorrefmark{2} Horizon Surgical Systems Inc.,Malibu, CA 90265} 
\IEEEauthorblockA{\IEEEauthorrefmark{3} Department of Radiological Sciences, University of California, Los Angeles, CA 90095 USA(email:dlu@mednet.ucla.edu;holdenwu@mednet.ucla.edu}}
\thanks{This work was supported in part by U.S. NIH Grant EB031934}}



\maketitle

\begin{abstract}
We present a MR safe, master-slave robot manipulator for abdominal interventions in the MRI chamber. A human operated 2+1-DoF master controller manipulator transmits motion and force to a 2+1-DoF slave manipulator via fluid transmission. Jointly, a digital master controller provides multimodal control capability beyond common split axis or mode switchable hybrid human-digital controller configurations found in previous studies. High input impedance, low-leakage, elastomeric fluid actuators are delegated to remote angulation control.  Low-friction graphite piston cylinders are delegated to needle insertion axis remote actuation given the sub-newton force transparency and sub-millimeter motion transmission over bedside fluid piping lengths. The device enables real-time MRI guided interventions allowing manual, digital, hybrid, and collaborative control modes.  Collaborative tasks such as assisted tissue penetration, fault-driven virtual fixture, and motion compensation through feedback control are presented in this paper. Preliminary MR scanner results demonstrate manipulator functional viability for an in-vivo pig experiment in bedside, manual control mode configuration.
\end{abstract}


\pagenumbering{arabic}
\setcounter{page}{1} 

\section{Introduction}
\IEEEPARstart{H}{epatic} and renal tumors represent a major threat to bodily health by potentially reducing the lifespan of both the liver and kidney. Magnetic resonance (MR) offers safe, high contrast, imaging of soft tissue inside the body.  In the context of tumor diagnosis and treatment, MRI has become a popular image guiding modality for percutaneous needle insertion procedures over computed tomography (CT) and ultrasound (US) \cite{Olthof2022-pb,Cazzato2021-zd,Cazzato2021-dr,Kerimaa2013-sv}. Percutaneous needle biopsy presents a host of challenges, including navigating the needle away from vital organs and blood vessels, ensuring precise puncturing, and achieving accurate targeted tissue positioning to guarantee effective treatment or accurate diagnosis.  \\
\begin{figure}[t]
    \centering    \includegraphics[width=1\linewidth]{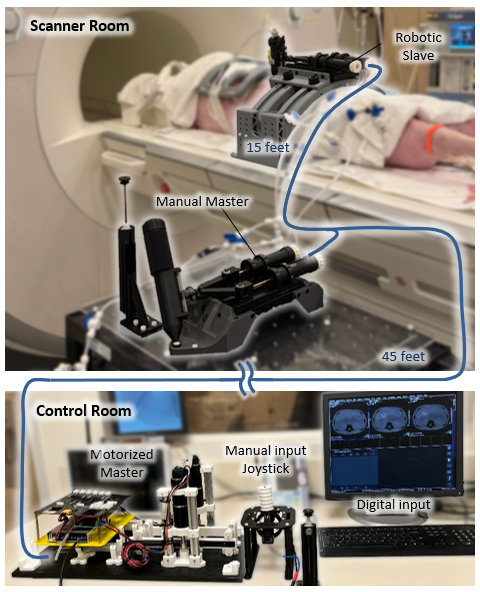}
    \caption{(Top) Manual master manipulator teleoperates the robotic slave mechanism in from the scanner room.  (Bottom) Digital master manipulator parallelly drives the robot slave from the control room. }
    \label{fig:MultiDOF_Pic1}
\end{figure} 
\indent Robotic needle insertion devices address the drawbacks related to accurate needle targeting.  In addition to reducing repositioning time, robotic mechanisms enable accuracy for instances where the needle is inserted at angles of $5^o$ or more from the axial/transversal plane \cite{Heerink2019-rn}. Unfortunately, obtaining quality images from MRI is subject to component compatibility requirements. Magnetic resonance imaging is a modality offering sharp image contrast for soft tissues to generate a three-dimensional online roadmap for surgical interventions.  An intense homogeneous static 3T magnetic field decays rapidly over a few meters from the bore's geometric center.  The high forces and torques on ferromagnetic or highly paramagnetic material in the scanner room presents a danger to the physician during the procedure.
Practical constraints have limited the realization of devices for in-bore MRI-guided surgical interventions. 
\\ 
\indent Needle artifact characteristics also present challenges to image quality consquently affecting accurate needle targeting \cite{Okamoto2021-jf,Hoffmann2016-mf,Lewin1996-qk,Salomonowitz2000-ma}. Furthermore, the protocol for many procedures is highly dependent on the skill of the surgeon, the size of the patient, and the location of the target tissue with respect to the skin layer.  Thus, this presents significant delays in the interventions as well introduction of targeting error that reduces the quality of the intervention and adds to the risk profile of surgery for the patient.  An assortment of image-guided surgical robots exist with each realization promising features unique to the surgical intervention type.

\begin{figure*}[t]
    \centering    
    \includegraphics[width=0.99\linewidth]{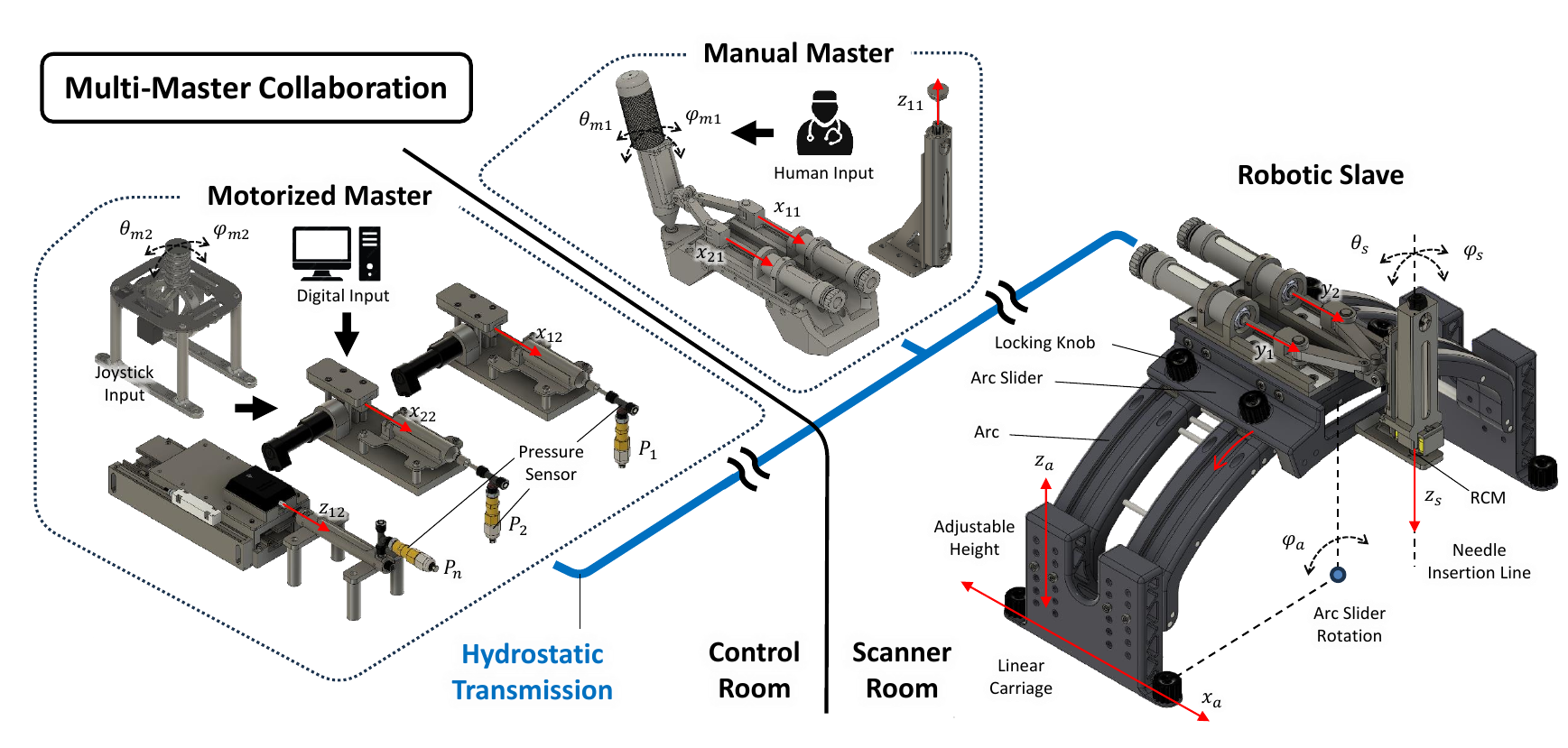}
    \caption{Multi-Master collaboration leverages motor-operated and human operated master angulation and insertion mechanisms for remote needle angulation and insertion via table-mounted hydrostatic, multi-DOF robot.}
    \label{fig:MultiDOF1}
\end{figure*} 
\subsection{Related Work}
\indent MRI-guided remote surgical robotics are prevalent in research for various applications ranging from needle biopsy, lesion ablation, and drug delivery to stereotactic neurosurgery,prostate surgery, and cardiac catheterization.
\\  
\indent Schreiber \textit{et al.} \cite{schreiber2022crane} presents a highly desterous, 10 DOF robot arm for CT guided percutaneous needle biopsy. Hata \textit{et al.} \cite{XYZold,XYZold2} presents a motorized remote-center-of-motion constrained needle guiding robot for MR-guided microwave thermotherapy.  Hata employed optical encoders on the end effector to track the accuracy of the RCM control.  During motorized RCM control, the electrical signal introduced significant noise in the image feedback. Morikawa \textit{et al.} \cite{MORIKAWA2009340} presents a remote-center-of-motion (RCM) control based MR compatible motorized manipulator for microwave coagulation therapy of liver tumors. Sato \textit{et al.} \cite{SatoI} presents an MR-guided, 2 DoF needle insertion manipulator system mounted on a 14 DoF ground based, positioning mechanism. Although ground based systems may offer a fixed, rigid reference frame, the bulk nature of these mechanisms occupies a large footprint making them only practically suitable for open-configuration MRI bores. \\
\indent Franco \textit{et al.} \cite{FrancoGantry,Franco-mr,Franco3DOF,FrancoPilotStudy} presents a table-mounted pneumatic, needle-guiding robot used for MR guided laser ablation of liver tumors along with a control strategy for pneumatic systems that improves force feedback in teleoperation\cite{FrancoMasterSlave,FrancoAdaptive,FrancoPredictive}. Zhang \textit{et al.} \cite{ZhangHybrid} propose a hybrid serial-parallel robot for percutaneous needle intervention procedures which consists of two layers of 3-RRR manipulators to position and orient the needle.  Meinhold \cite{Meinhold} achieves sub-pixel positioning accuracy with a parallel plane MRI needle guide actuated by piezoelectric motors. Moreira \cite{Miriam} uses a combination of pneumatic and piezoelectric actuation to drive a 5 DOF parallel needle guide and a 4 DOF needle insertion module. Su \textit{et al.} \cite{Su1} presents a similar prostate therapy, MRI-guided, 6-DOF needle driver module. Liang \textit{et al.} \cite{LiangDoubleArch} presents a table-mounted MR safe, 5 DoF, parallel double-arch needle insertion mechanism with a scissor folding mechanism. The arch is situated on a pneuamtically driven linear gantry while the scissor mechanism is actuated by a pneumatically driven timing belt mechanism.  Unfortunately, the pneumatic actuators require reliable and high working pressures to deliver sufficient torque to drive the belt drive.  The timing belts generally suffer from nonlinear friction which may result in transient motion artifacts.  Though table-mounting mechanism meet space constraint criteria for closed-MR bore applications, most of the mechanisms rely on pneumatic, piezoelectric, or ultrasonic motors for actuation.  Pneumatic systems require precise pressure control valve components that must remain a significant distance from the bore.  Piezoelectric components, although MR-conditionally safe, can potentially introduce noise into the MR image feedback. \\
\indent Li and Wang \textit{et al.} \cite{Li2020-rj,InsituLumbar,InsituLumbar2,InsituLumbar3} develop a body-mounted, MR safe, 6 DOF robot equipped with a 2-DOF remotely actuated needle driver module using beaded-chain transmission for lumbar spinal injection.  He \textit{et al.} presents soft fluid actuator driven, body mounted, 2 DoF cross arch needle orientation mechanism for percutaneous ablation in hepatocellular carcinoma \cite{Zhuoliang1}. Yan and Patel \cite{BodyMounted} present a MRI/CT compatible 4-DOF parallel robot for shoulder arthrography guided needle-based percutaneous interventions.  Beaded chain transmission mechanism overcome the MR safe requirements for remote actuation mechanisms but require low manufacturing tolerance and precision tension to ensure reliable motion transmission. Though body-mounted mechanism offer a more flexible and adaptable solution, the viable needle workspace may be limited given the size and weight constraint of the robot. Furthermore, body-mounted mechanisms suffer from net displacement due to diaphragm motion during breathing. 
\indent Lee \textit{et al.} \cite{InterCardiacCath} introduces an MR compatible, multi-DOF intercardiac catheterization hydraulic, rolling diaphragm robot that employs an electric motor driven master-slave actuation of rolling diaphragm actuators. Guo \textit{et al.} \cite{StereoNeuro} presents an MRI-guided robot capable of performing bilateral neuro-sterotaxy for the ellipsoid like subthalamic nucleus (STN) about 90 mm beneath the skull.  Although hyrdraulic based system offer improved power throughput compared to pneuamtic systems they still require external motors to provide power.  This may result in heavier ferrous machinery placed near the MRI bore. \\
\indent Hydrostatic rolling diaphragm actuators are commonly adopted in medical applications primarily for their power-dense actuation.  They overcome the stick-slip friction effects from sliding seals in hydraulic pistons. Frishman \textit{et al.} \cite{Frishman7DOF,Frishman7DOF_2} presents a passive,  hydrostatic teleoperator consisting of a MR safe, six-axis arm and needle insertion end effector.  The teleoperator is designed to be backdrivable and force transparent.  Force transparency is presented as a feature in the robot to reduce the membrane overshoot when puncturing through tissue layers.  MR-safe rolling-diaphragm linear actuators are popularly adopted for prostate needle biopsy \cite{Evelyn}, and human-friendly robotic arms \cite{Lightweight,Supernumerary,LowLevel}.  Long-stroke fluid actuators have been explored \cite{Hashemi1} but commercially available rolling diaphragms are limited to a stroke-to-diameter ratio of one.  This usually demands the incorporation of linkages, belts, or cables to magnify the motion for needle robots. Hence \cite{Frishman7DOF} presents a six-axis serial manipulator to meet the achievable workspace criteria necessary for abdominal interventions.   

\begin{table}
\centering
\begin{tabular}{lllll} 
\toprule
 \textbf{Ref} & \textbf{MRI-SL} & \textbf{F-F} & \textbf{Mm-C}& \textbf{RMC}
 \\
 \midrule 
  \cite{Zhuoliang1} &  Safe  & No & Hybrid & No\\
 
  \cite{schreiber2022crane}  &  Unsafe& No & Digital & No\\
  
  \cite{XYZold,XYZold2} &  Conditional & No & Hybrid & No \\

  \cite{SatoI} &  Conditional& No & Hybrid & No\\
 
    \cite{FrancoGantry,Franco-mr,Franco3DOF,FrancoPilotStudy,FrancoAdaptive,FrancoMasterSlave,FrancoPredictive} &  Conditional& No & Hybrid & No \\ 

   \cite{ZhangHybrid} &  Unsafe & No  & Digital & No \\

   \cite{Meinhold} &  Conditional& No &  Hybrid & No\\

  \cite{Miriam} &  Conditional  & No & Digital & No\\
   
   \cite{Su1} &  Conditional  & No & Digital & No\\ 

    \cite{LiangDoubleArch} &  Safe  & No & Hybrid & No \\
   
   \cite{Li2020-rj,InsituLumbar,InsituLumbar2,InsituLumbar3} &  Conditional & No & Hybrid & No\\
   
   \cite{BodyMounted} &  Conditional & No & Hybrid & No \\
   
   \cite{InterCardiacCath} &  Safe & No & Hybrid & No \\ 
   
   \cite{Frishman7DOF,Frishman7DOF_2} &  Safe & Yes & Manual & No\\
   This paper &  Safe & Yes & Hybrid\&Collab.& Yes \\
   \bottomrule
 \end{tabular}
\caption{Remote needle insertion robots for abdominal interventions qualified by their criteria satisfaction level. MRI safety level follows the standard \cite{ASTM_F2503_23E01}. \emph{Manual} --- the operator's
own hand motion, transmitted directly or through a passive link; \emph{Digital} --- motors
execute the motion, the operator supplies targets or commands; \emph{Hybrid} --- either manual or digital on
different axes or switchable on same axis. \textbf{Collaborative} mixing of manual and digital inputs on the same axis. \textbf{MR-SL}: MR safety level. \textbf{F-F}: Force feedback. 
\textbf{Mm-C}: Multi-mode control \textbf{RMC}: Respiratory motion compensation. \label{tab:LitTable3}}
\end{table}

\subsection{Contribution}
\indent The aim of this paper is to present a table-mounted remote, MRI safe, needle insertion robot that operates in closed-bore MRI scanners, offers significant reduction in procedure duration, and achieves remote \emph{in-situ} needle position and orientation control.  The robot is mounted on a circular arch that fits in the 20 cm gap between the patient skin and the inner walls of a  60 cm bore.  Fromt the scanner room, the robot mechanism consist of a hydrostatically actuated 2 DoF needle, P-RRR-P serial-parallel hybrid, angulation mechanism.  The actuators are long stroke ($>4 \text{in}$) elastomer sealed piston cylinders. To allow \textit{in situ} needle angle adjustment, a bedside, hydrostatic, manual master joystick remotely drives the needle angle.  A single DoF needle insertion master-slave hydrostatic cylinder pair bi-directionally drives the needle into and out of the target tissue.  The insertion module consists of a low-friction graphite piston actuator to enable force feedback for needle-tissue interaction sensing. 
From the control room, the robotic system is augmented with a  collaborative digital control by a motorized master unit allowing for direct, programmatic  control commands by digital user-interface or via a physical, encoder-based joystick module.  The resulting fluid volume inputs are combined at the fluid network level to accomplish collaborative tasks such as those detailed in section \ref{sec:EvalResults}.

The needle insertion module is mounted on a semi-circular arch to maximize workspace and allow oblique needle insertion angles necessary for environmentally-sensitive needle trajectories.  In contrast to the active/passive RCM designs from previous studies, this robot mechanism uses a fixed center-of-motion by a spherical pivot that is fixed and tangent to the pre-insertion,skin entry point. Recent studies involving hydrostatically actuated liver surgery needle angulation mechanisms \cite{Frishman7DOF,Evelyn} conclude that back-drivability is a desirable attribute for abdominal interventions. This robot shows exceptional force transparency along the needle axis down to $\pm 0.5$ N range.  The robot is manufactured with MR-safe materials to guarantee no electromagnetic coupling with the MR scanner and no induced image artifacts.  \\
\indent This paper is organized as follows. Section \ref{sec:SysReqWork} presents the system's target requirements for abdominal surgical interventions.   Section \ref{sec:RobSys} presents the robot mechanics, forward and inverse kinematics, and the needle's reachable workspace. Section \ref{sec:EvalResults} presents the evaluation and results for the robot's static and dynamic characteristics. Lastly, in section \ref{sec:HILNM} an MR image sequence for a \textit{in situ} needle targeting procedure is presented to corroborate the robot's viability in clinical applications. 

\section{System Requirements} \label{sec:SysReqWork}
\subsection{System Requirements}
\noindent The requirements for the robot are listed as follows: 
\begin{enumerate}
    \item \emph{MRI Safe and lesion reachability} 
    \item \emph{Force feedback} 
    \item \emph{Multimode control}
\end{enumerate}

\subsubsection{MRI Safety and lesion reachability} The system should be MR safe [ASTM F2503]\cite{ASTM_F2503_23E01} by minimizing magnetically induced forces, induced voltages, and minimizing heating on conductive materials. In addition, the robot should not induce image distortion artifacts or significant reduction in the signal-to-noise (SNR) ratio within the region of interest (ROI). To avoid this, the slave manpulator is free of any local optical or electrical components such as rotary or linear incremental encoders. \\
\indent
The table-mounted robot fits in closed-MRI scanner ranges 60 to 70 cm \cite{Patel2019-so} for gantry to skin distance of 10-20 cm \cite{Seimenis2012-ns,Bricault}. The robot workspace contains the majority of the liver volume for a patient in the supine position.  This implies that the angular insertion range should be $\pm 45^o$ from perpendicular to skin \cite{Franco-mr}. Additionally, some clinical interventions require oblique needle insertion paths to avoid healthy organs or bones, therefore, requiring the follower robot to have an additional $\pm$ 20 degree gross arch angle range. Target lesions should be reachable with the needle. 
\\
 

\indent Hepatic tumors mean depth is about $5$ cm with a maximum of $15$ cm \cite{Heerink2019-rn,Ben-David2018-iq,Yamanaka2015-rj}. The dimensions are parameterized in three diameters, AP-anteroposterior diameter, LL-laterolateral diameter, and the CC-craniocaudal diameter with  average lengths being 14.5 cm of AP, 18.5 cm for LL, and 15.5 cm for CCmax diameters \cite{drvendzija2023liver,nagato}.  Given the liver dimensions and mean depth, the needle tip's workspace should intersect with at least $33\%$ of the liver volume.
\\
\subsubsection{Force Feedback} The robot manipulator should have high motion and force responsiveness to operator input. 


The requirement for this robot is to have a back-drivable transmission to allow finer mechanical impedance profile over long fluid transmission lines. Hydrostatic actuation offers operator haptic feedback to assist surgeons by providing guidance by tactile cues in addition to the image feedback. \\
\subsubsection{Multimode Control} The master-slave robot manipulator should be multi-master compatible.  The robot leverages multiple master inputs summed in the fluid volume domain for automated and semi-automated procedures.  This implies that the needle angulation and insertion module can offer a collaborative master configuration that can assist the surgeon in needle angulation and insertion, particularly under the presence of needle loading changes as it advances through the tissue.

\section{Robot Mechanics} \label{sec:RobSys}

\subsection{Remote Needle Insertion Robot}
The remote needle insertion mechanism consists of a four-bar linkage mechanism with a translating and rotating base link and with the output to coupler joint at a fixed angle.  Instead of driving the input link with a prescribed torque input, the mechanism self aligns according to the displacement of the revolute joints as the base link.  The resulting mechanism can be classified as a P-RRR-P mechanism with two prismatic joints and three revolute joint.  The mechanism is considered a hybrid serial-parallel manipulator due to the 4-bar linkage responsible for the differential and translational mode that results in 2-DoF angulation of the needle.

\begin{table}[h!]
\centering
\begin{tabular}{ cccccc } 
 \toprule
 \textbf{Parameters} & \textbf{Description} & \textbf{Value (units)} \\
 \midrule
$R_a$ &Radius of arch &30 (cm) \\ 

$L_s$ &Rail length of cylinders &10.8 (cm) \\ 

$R_s$ &Radius of cylinder pistons &1.14 (cm) \\ 

$A_s$ & Area of cylinder pistons & 4.1 ($\text{cm}^2$)\\

$L_w$ & length of tubing & 15-45 (ft) [4.6-13.7 (m)] \\ 

$L_n$ & length of needle cylinder & 11.6 (cm) \\

$d_p$ & diameter of tubing & ID=4 (mm), OD=6.35 (mm) \\

$l_p$ & length of needle piston & 1.3 (mm) \\

$A_0$ & Area of needle piston & 0.675 ($\text{cm}^2$)\\
\bottomrule
\end{tabular}
\caption{Robot Manipulator nominal dimensions for closed 60 cm MR bores as defined in figure  \ref{fig:MultiDOF1}.\label{tab:FirstTable}}
\end{table}

The robot manipulator is designed for 40-70 kg in-vivo pig subjects in 60 cm, closed MRI bores but can be readily scaled to meet the human-sized 70 cm MR bores for  clinical applications. 

\subsection{Slave Needle Tip Workspace}
The robot end effector requires particular workspace requirements to allow the leader manipulator to adjust the needle tip during tissue targeting. Due to the high impedance of the elastomeric fluid actuators, the remote needle insertion can be performed at any non-zero needle angle with respect to gravity. During liver biopsy the needle tip is expected to reach a target at about 10 cm below the incision point.  The needle angle can range between -40 to 21 degrees along the sagittal plane and -65.5 to 50 degrees along the axial plane with respect to gravity.

\begin{figure*}
\centering
\subfloat[]{\includegraphics[width=0.45\textwidth]{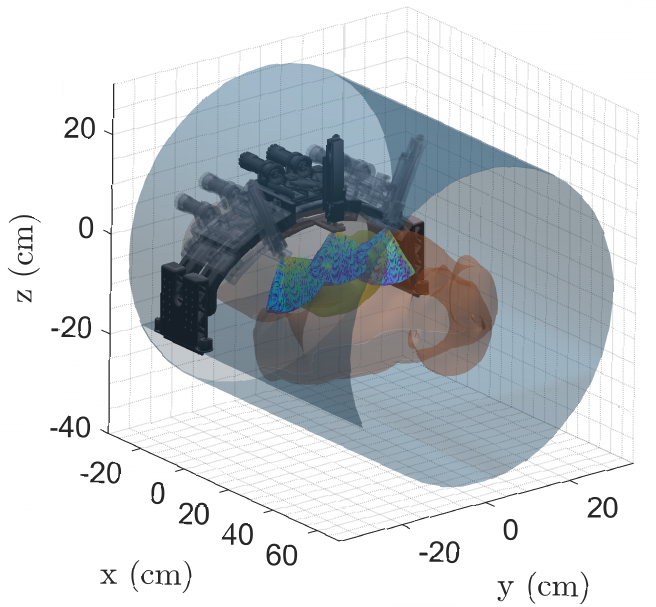}\label{fig:sub1}}\hskip 1ex
\subfloat[]{\includegraphics[width=0.45\textwidth]{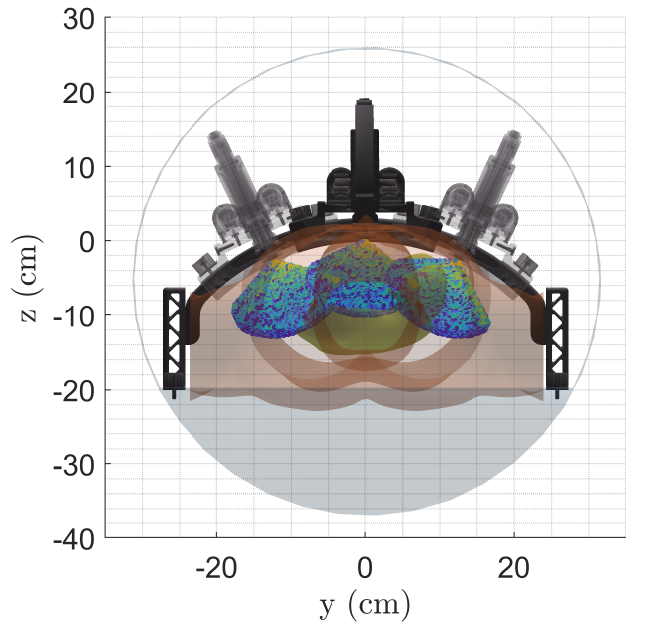}\label{fig:sub2}}
\caption{ (\ref{fig:sub1}) 
Isometric view of scaled patient inside 60-cm bore with 3 overlayed angular positions, $\varphi_a=-20,0,20$ degrees. 
(\ref{fig:sub2}) Axial view of robot manipulator tool tip density map overlayed for 3 fixed arch, angular positions. }
\label{fig:RobotWorkSpace}
\end{figure*}

The density map for the needle tip position is shown in figure \ref{fig:sub2}. The density is defined a natural logarithm of the number of possible configurations over the maximum number of possible configurations over the set of all locations. 

The resolution of the workspace analysis is $1mm$. Note that the density map is heavily weighted in the region near the pivot arm but loses versatility for regions nearing the extremes of the bounding box.  
Using \cite{drvendzija2023liver,nagato}, the robot's reachable workspace intersects with $33.3-50\%$ of the the liver volume assuming it is 10 cm from the skin's surface with a anteroposterior diameter of about 15cm and a needle length of 12 cm.  If, the needle insertion plan requires an oblique entry path, then the total intersected volume would be reduced. 

\subsection{Robot Kinematics: 4-bar linkage} 

The modified-DH parameters are written for the constrained 4-bar linkage in tables the left and right sub-tables denoted in table \ref{tab:FullDHTable}.
The robot manipulator mechanism can be decomposed into 2 modules. The first module is the pair of linear actuators that displace the spherical joint that holds the needle insertion module via a set of linear guide rails.  This segment acts as a under-constrained 4-bar linkage.  However, the linkage becomes fully constrained by fixing the input link angle and actively changing the base plate's length.  The resulting kinematics propagate to the needle insertion module which is pinned to a spherical joint at the pivot arm. 


\subsection{Robot Kinematics: Needle Angulation Pivot Arm }

\begin{figure}
    \centering
    \includegraphics[width=0.99\linewidth]{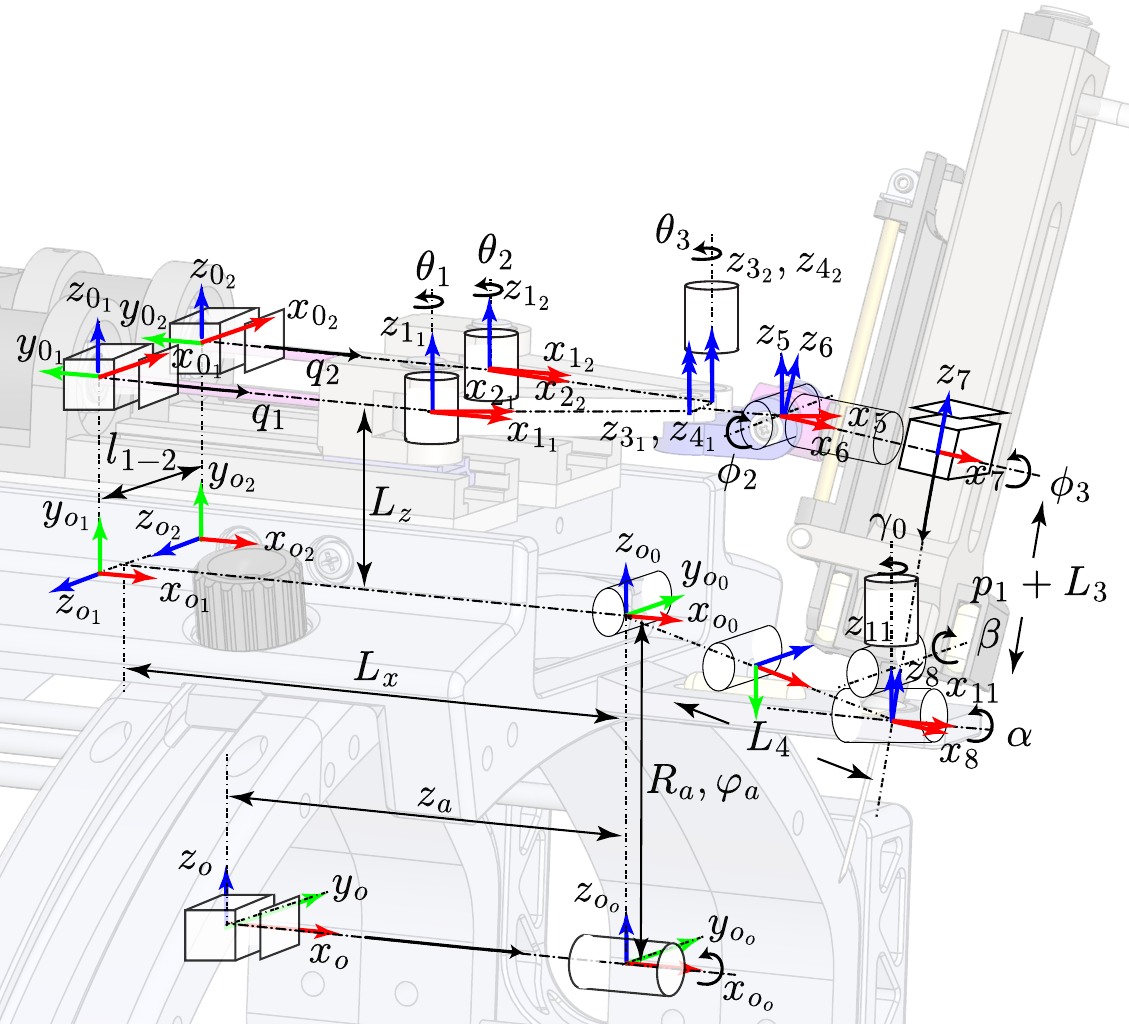}
    \caption{Kinematic diagram of serial-parallel slave needle insertion robot with associated modified-DH table in table \ref{tab:FullDHTable}.}
    \label{fig:KinematicsFrames}
\end{figure}

The robot kinematics governed by the 4-bar linkage drive the needle insertion module \textcolor{blue}{handle}.  The origin lies near the pivot arm.  There are two serial paths for defining the configuration of the needle module's body frame.  The serial paths can be defined by a series of homogeneous transformations that can be defined by appropriated modified DH parameter table. 
The first two frames in the full DH table denote the translation of the arch on the MR bed ($z_a$) and the rotation of the manipulator mount along the arch ($\varphi_a$). 
From frame $o_0$ the forward kinematics are split into two paths: one for the right actuator and the other one for the left actuator. The transformations meet on frame 5 to and continue down to the pivot point before following the path back towards frame $o_0$.  Figure \ref{fig:KinematicsFrames} expresses the needle's angle with a standard XYZ rotation sequence ${\alpha,\beta,\gamma_0}$.  A conventional parameterization is the azimuthal ($\theta$) and elevation ($\varphi$) which follows for the remainder of this paper.

\begin{table*}[]
    \centering
    \begin{tabular}{|c|c|c|c|c|c|c|c|c|c|c|c|c|}
        \hline 
        \multicolumn{2}{|c|}{\textbf{Frame}} & \multicolumn{2}{c|}{\textbf{Type}} & \multicolumn{2}{c|}{$\boldsymbol{\alpha_{i-1}}$} & \multicolumn{2}{c|}{$\boldsymbol{a_{i-1}}$} & \multicolumn{2}{c|}{$\boldsymbol{d_i}$} & \multicolumn{2}{c|}{$\boldsymbol{\theta_i}$}  \\
        \hline 
        \hline 
        \multicolumn{2}{|c|}{$o_o$} & \multicolumn{2}{c|}{p} & \multicolumn{2}{c|}{0} & \multicolumn{2}{c|}{$z_a$} & \multicolumn{2}{c|}{0} & \multicolumn{2}{c|}{0} \\
        \hline 
        \multicolumn{2}{|c|}{$o_0$} & \multicolumn{2}{c|}{r} & \multicolumn{2}{c|}{$\varphi_a$} & \multicolumn{2}{c|}{0} & \multicolumn{2}{c|}{$R_a$} & \multicolumn{2}{c|}{0} \\
        \hline
        \hline 
        \textbf{Frame}[1] & \textbf{Type}[1] & $\boldsymbol{\alpha_{i-1}}$ [1] & $\boldsymbol{a_{i-1}}$ [1] & $\boldsymbol{d_i}$ [1] & $\boldsymbol{\theta_i}$ [1] & \textbf{Frame}[2] & \textbf{Type}[2] & $\boldsymbol{\alpha_{i-1}}$ [2] & $\boldsymbol{a_{i-1}}$ [2] & $\boldsymbol{d_i}$ [2] & $\boldsymbol{\theta_i}$ [2]
        \\
        \hline 
        $o_1$ & o & $90$ & $-L_x$ & $l_{1-2}/2$ & 0 & $o_2$ & o & $90$ & $-L_x$ & $-l_{1-2}/2$ & 0 \\
        \hline 
        $0_1$ & o & $-90$ & $0$ & $L_z$ & $90$ & $0_2$ & o & $-90$ & 0 & $L_z$ & 90 \\
        \hline 
        $1_1$ & p & $90$ & 0 & $q_1$ & 0 & $1_2$ & p & 90 & 0 & $q_2$ & 0 \\ 
        \hline 
        $2_1$ & r & $-90$ & 0 & 0 & $-90+\theta_1$ & $2_2$ & r & $-90$ & 0 & 0 & $-90-\theta_2$ \\
        \hline 
        $3_1$ & r & 0 & $l_1$ & 0 & $\varphi_1$ & $3_2$ & r & 0 & $l_2$ & 0 & $-\theta_3$ \\
        \hline 
        $4_1$ & r & 0 &$d_1/2$ & 0 & $-90$ &
        $4_2$ & r & 0 &$d_1/2$ & 0 & $90$ \\
        \hline 
        \hline 
        \multicolumn{2}{|c|}{$5$} & \multicolumn{2}{c|}{r} & \multicolumn{2}{c|}{0} & \multicolumn{2}{c|}{$D_1$} & \multicolumn{2}{c|}{0} & \multicolumn{2}{c|}{0} \\
        \hline 
        \multicolumn{2}{|c|}{$6$} & \multicolumn{2}{c|}{r} & \multicolumn{2}{c|}{$-90$} & \multicolumn{2}{c|}{0} & \multicolumn{2}{c|}{0} & \multicolumn{2}{c|}{$\phi_2$} \\
        \hline 
        \multicolumn{2}{|c|}{$7$} & \multicolumn{2}{c|}{r} & \multicolumn{2}{c|}{$90+\phi_3$} & \multicolumn{2}{c|}{$L_1+L_2$} & \multicolumn{2}{c|}{0} & \multicolumn{2}{c|}{0} \\
        \hline
        \multicolumn{2}{|c|}{$8$} & \multicolumn{2}{c|}{p} & \multicolumn{2}{c|}{0} & \multicolumn{2}{c|}{0} & \multicolumn{2}{c|}{$-p_1-L_3$} & \multicolumn{2}{c|}{0} \\
        \hline 
        \multicolumn{2}{|c|}{$9$} & \multicolumn{2}{c|}{r} & \multicolumn{2}{c|}{180} & \multicolumn{2}{c|}{0} & \multicolumn{2}{c|}{0} & \multicolumn{2}{c|}{$\gamma_0$} \\
        \hline
        \multicolumn{2}{|c|}{$10$} & \multicolumn{2}{c|}{r} & \multicolumn{2}{c|}{$-90$} & \multicolumn{2}{c|}{0} & \multicolumn{2}{c|}{0} & \multicolumn{2}{c|}{$\beta$} \\
        \hline
        \multicolumn{2}{|c|}{$11$} & \multicolumn{2}{c|}{r} & \multicolumn{2}{c|}{$-90-\alpha$} & \multicolumn{2}{c|}{0} & \multicolumn{2}{c|}{0} & \multicolumn{2}{c|}{0} \\
        \hline 
        \multicolumn{2}{|c|}{$12$} & \multicolumn{2}{c|}{r} & \multicolumn{2}{c|}{$-90$} & \multicolumn{2}{c|}{0} & \multicolumn{2}{c|}{0} & \multicolumn{2}{c|}{$\phi_4$} \\
        \hline 
        \multicolumn{2}{|c|}{$13$} & \multicolumn{2}{c|}{r} & \multicolumn{2}{c|}{0} & \multicolumn{2}{c|}{$-L_4$} & \multicolumn{2}{c|}{0} & \multicolumn{2}{c|}{$-\phi_4$} \\
        \hline 
        \multicolumn{2}{|c|}{$o_0/14$} & \multicolumn{2}{c|}{r} & \multicolumn{2}{c|}{$90$} & \multicolumn{2}{c|}{0} & \multicolumn{2}{c|}{0} & \multicolumn{2}{c|}{0} \\
        \hline 
    \end{tabular}
    \caption{Full modified-DH parameter table for the hybrid, serial-parallel master-slave needle insertion manipulator. The first 2 DoF describe the translation and rotation about the arch.  The path splits into 2 to model the constrained 4-bar linkage before combing at frame 5. Frame 5-14 is consistent with a singular serial chain.}
    \label{tab:FullDHTable}
\end{table*}

\subsection{Inverse Kinematics}

The inverse kinematics involves solving for the fluid actuator displacements, $q_1$ and $q_2$ as a function of the desired needle angle in terms of the azimuthal ($\theta$) and elevation ($\varphi$).  If the angle sequence is in a 3-axis Euler angle sequence, then, an iterative solver \ref{alg:cap} is necessary to determine the 3rd angle. Conventionally, the needle command can be parameterized as a quaternion or rotation sequence which can be inverted using closed-form inverse kinematics techniques to determine the actuator displacements.

\begin{figure*}[h!]
    \centering
    \includegraphics[width=0.995\linewidth]{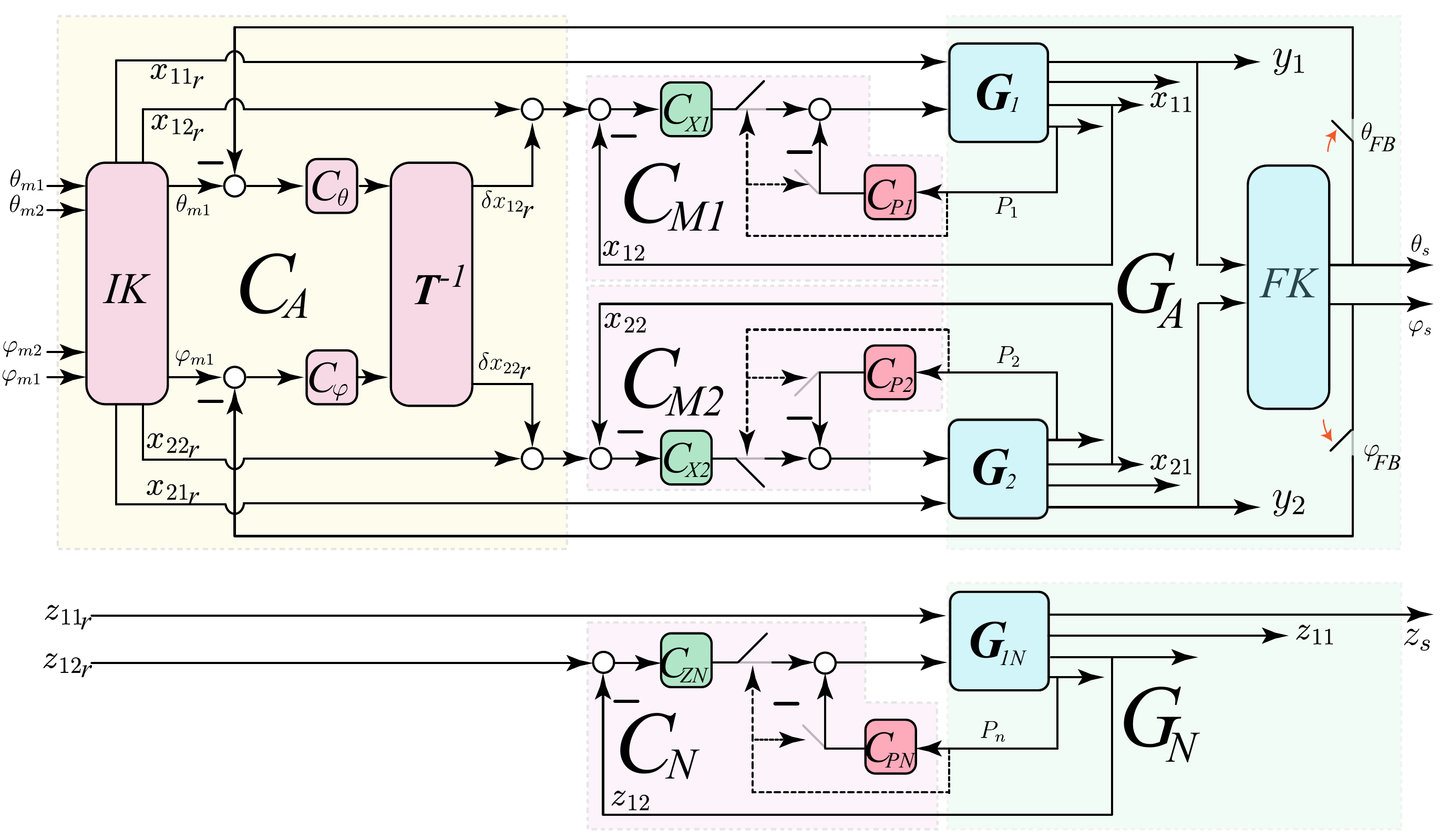}
    \caption{Control architecture for master-slave teleoperated needle robot includes internal servo-mechanisms ($C_{Mi},C_{N}$) for fluid actuator fault-triggered pressure regulation, ($C_{Pi}$), and motion tracking, $C_{Xi},C_{ZN}$. Similarly, an output angle regulation controller ($C_A$) contains motion compensation and coordinate transformation, ($C_{\theta},C_{\varphi},IK,T^{-1}$). }
    \label{fig:SystemBlockDiagramv1}
\end{figure*}

\section{MR Scan Robot Characterization}
\subsection{MR Compatibility}
To quantify the signal to noise ratio (SNR) for the MR image feedback we perform a scan for a matrix of robot configurations that include robot near but outside the MR bore as well as inside the bore. The images were acquired using a spin echo sequence (TR/TE = 500/20ms, flip angle=90°, FOV=250mmx250mm, spatial resolution= 0.98mmx0.98mm).  Results a $0.11\%$ difference in SNR between the nominal MR images and those acquired with a bedside robot. 
Furthermore, while the robot is inside the bore it demonstrates a $0.59\%$ difference in SNR.  Additionally, distortion effects are also observed to be $(0.2,0.2)\%$ difference with the bedside robot along the sagittal and axial axes while a $(0.52,0.2)\%$ difference with an in-bore placement of the robot.  These findings confirm the satisfaction of the standardized MR safety level requirement for MR safe devices. Furthermore, the image feedback obtained from the scanner is minimally affected ensuring clear and identifiable objects in the image stream.

\subsection{Robot Workspace MR Scans}
To validate the robot needle tip workspace volume, a set of 10 static needle poses were scanned and process to validate the range of the kinematically-derived workspace of section \ref{sec:RobSys}.  The needle was manually detected on the image and the centroid of the signature was used to determine it's angle.  The corresponding result was then projected onto a 10 cm long needle and confirmed to lie inside the convex hull of the model's reachable workspace. 

\section{Evaluation and Results} \label{sec:EvalResults}

\begin{figure*}
    \centering
    \includegraphics[width=0.99\linewidth]{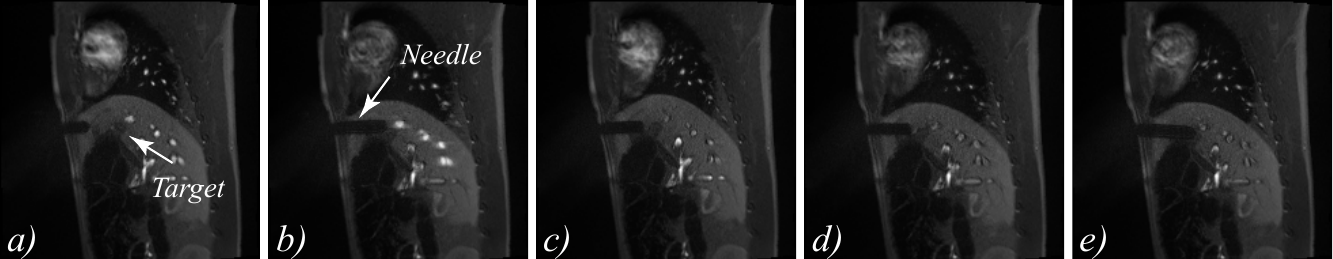}
    \caption{a) Needle is inserted into pig about 5-7 cm depth from the skin. b) Needle is slowly advanced forward as it nears the target. Needle reaches target depth and image clearly shows it missed the target's centroid. c) Needle is retracted slightly and angle is adjusted \textit{in-situ} to coincide needle tip with target centroid. (d) Needle angle is adjusted to realign with target path. (e) Needle advances and reaches the target centroid.}
    \label{fig:PigExpFinal}
\end{figure*}

\begin{figure*}[h!]
\centering
\subfloat[]{\includegraphics[width=0.305\textwidth]{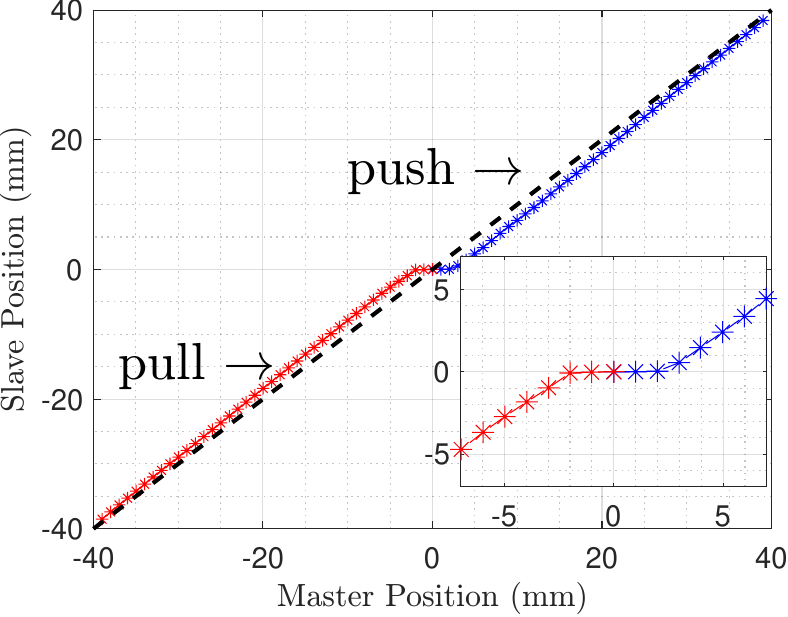}\label{fig:step40mm}}\hskip1ex
\subfloat[]{\includegraphics[width=0.305\textwidth]{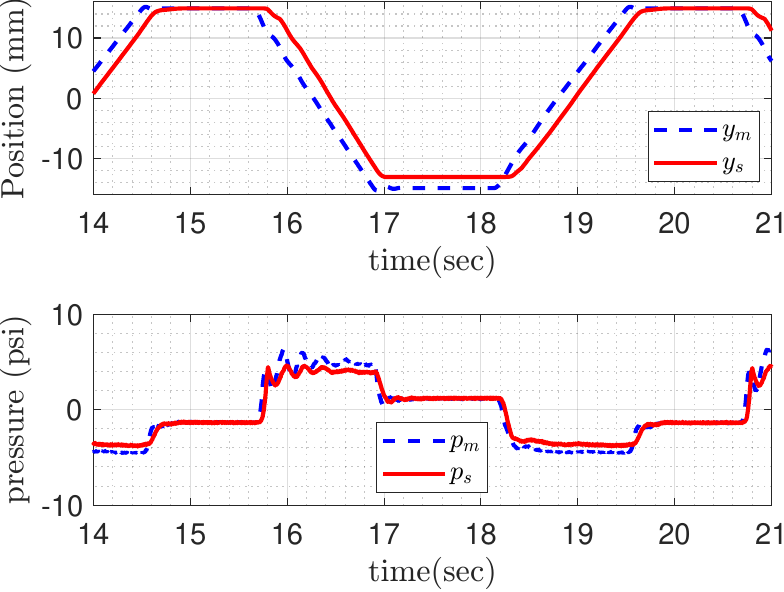}\label{fig:trap5sec}}
\subfloat[]{\includegraphics[width=0.305\textwidth]{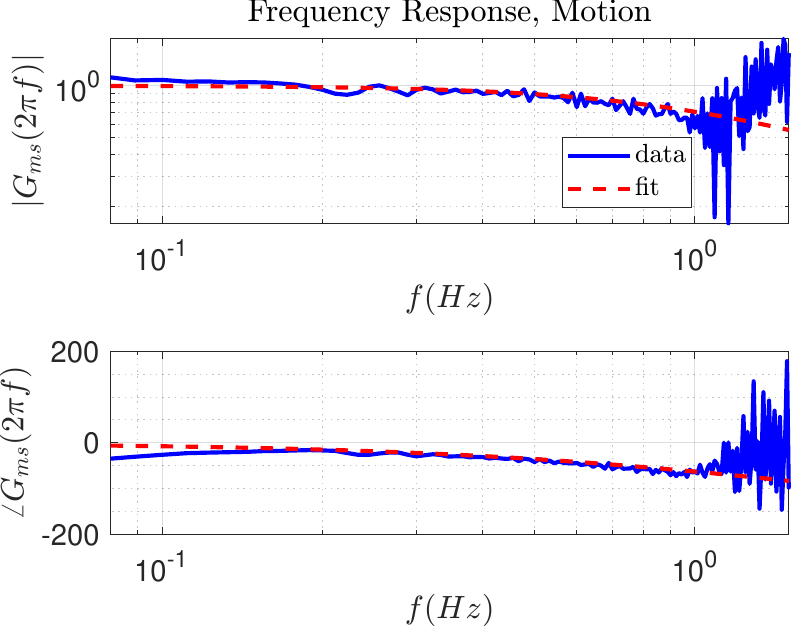}\label{fig:FreqrespMot}}
\caption{\ref{fig:step40mm} The motion transmission linear correlation of $0.945$ from master to slave for a range of 40 mm in the compression direction and $0.975$ in the tension direction.\ref{fig:trap5sec} Dynamic motion transmission of master to slave for a $30$mm pk-pk, T=$5$sec trapezoidal waveform.  The corresponding water pressure transmission in subplot B.\ref{fig:FreqrespMot} Frequency response of angulation fluid actuators generated by a band-limited chirp signal $f\in [0.1,1]Hz$. A single-pole filter with 0.1 second delay was used to fit the data.}
\end{figure*}

Fluid actuators are differentiated by axis. Most low-friction sliding-piston actuators suffer from fluid leakage during loaded retraction due to internal pressure reaching less than 1 atm. Fluid leakage result in force transparency and motion transmission loss due to stiffness loss by air entrapment. For actuators used in the angulation module of both master and slave, an elastomeric piston cylinder provides the necessary sealing specifications to meet load demands of the mechanism.  
\indent Although the piston seal properties may be advantageous for continuous cycling and re-positioning, the drawback is a loss in force transparency to the introduction of piston friction as it compresses against the inner walls of the cylinder. Abdominal surgical interventions rely on force feedback as a guide when readjusting needle angles or insertion depths. Given the minimal pulling forces from the tissue on the needle axis, a low-friction glass cylinder is used to drive and retract the needle during procedure.

\subsection{Force Feedback}
\subsubsection{Elastomeric Fluid Actuators}
The use of elastomeric-sealed fluid actuators allows for high impedance motion transmission due to the inherent friction in the piston.  The trade-off sacrifice is a small deadzone \ref{fig:step40mm} for small displacements which can be alleviated by feedforward high-frequency dithering superimposed onto the reference waveform.

The perform this motion transmission evaluation, the fluid actuator piston rods are screwed into a rack gear that connects to a pinion gear with the appropriate constraints.  The mechanism is similar to the hydrostatic motion transmission system described in \cite{Simonelli1}.  The pinion gear is mounted onto the motor shaft via a flexible, self-aligning coupler.  

\indent A set of reference waveforms were used to evaluate different characteristics of the fluid actuator.  Observing figure \ref{fig:step40mm}, the experiment involved tracking a bidirectional staircase waveform.  The staircase consists of 40 mm total displacement split into 40 1 mm steps in the forward (compression) and reverse (tension) directions. To characterize the motion transmission from master to slave, 1-1 line is used as comparison. A small dead-zone of $2.5$ mm is identifiable from the data illuminating the expected drawback from the elastomeric fluid actuators.  

Due to the high stiction in the actuators, there is a propensity for the pressure in the fluid line to vary among the steady-state intervals of the waveform.  There is also friction variability along the stroke of the actuator due to irregularities in the surface finish of the cylinder's interior walls.  From the pressure feedback, it was observed that, during compression, the fluid pressure fluctuates around $5$ psi ($P_i A_s =14.2$ N) but then settles at $2.3$ psi ($P_i A_s =5.7$N) implying that the angulation mechanism has a holding force of $\approx 6$ N. This follows that the stiction in the actuator is expected to be around 6-8 N.  The motion transmission as seen in figure \ref{fig:step40mm} demonstrates that the accumulated pressure from the dead-zone is progressively relieved as the motion extends to its maximum displacement.  Lastly, a sine-sweep experiment demonstrates a motion transmission bandwidth of 1 Hz accompanied by a 100 ms transmission delay due the fluid network tubing length of 40 ft.

\subsubsection{Low-friction needle insertion actuators}

\begin{figure}[h!]
\centering
\subfloat[]{\includegraphics[width=0.425\textwidth]{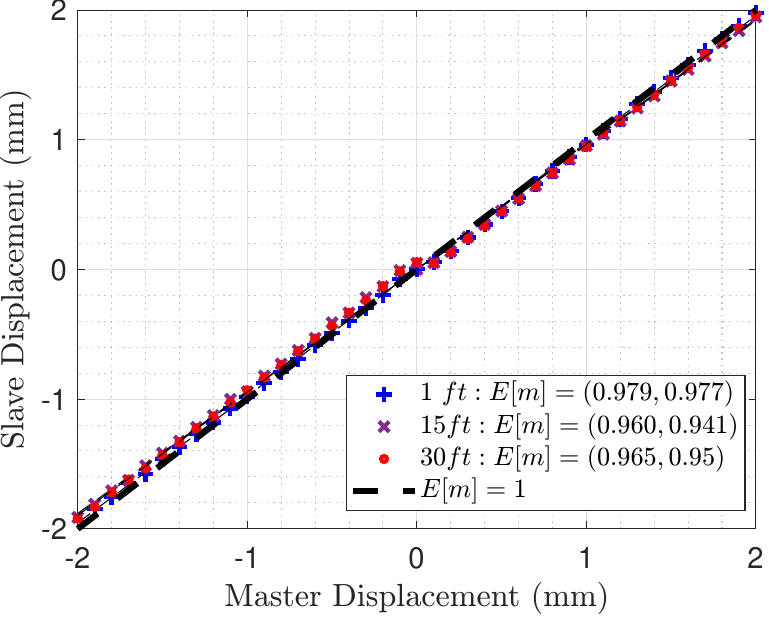}\label{fig:GraphitePistonMotion}} \hskip 1ex
\subfloat[]{\includegraphics[width=0.425\textwidth]{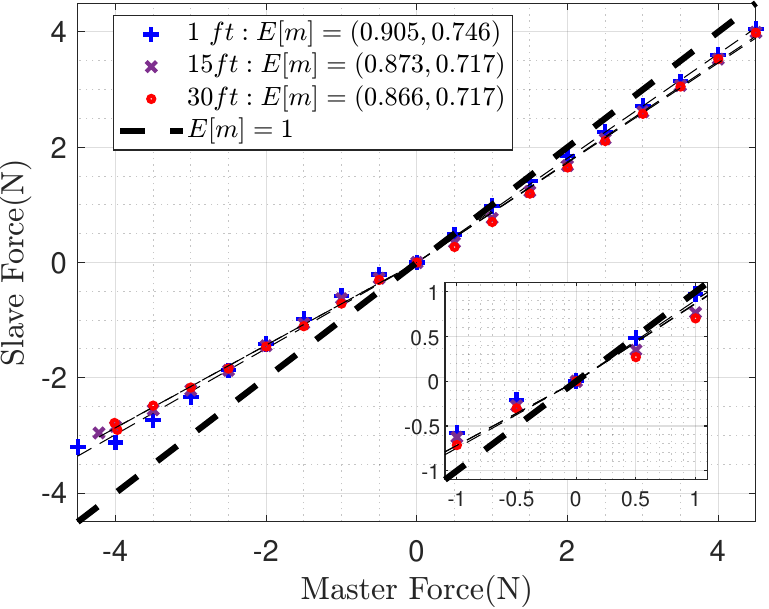}\label{fig:GraphitePistForce}}
\caption{\ref{fig:GraphitePistonMotion} Shows the correlation between master and slave load-less motion transmission for a range of $\pm 40$ mm. \ref{fig:GraphitePistForce} shows the linear master-slave force correlation for the \emph{needle insertion module} in the push and pull directions.}
\end{figure}

Low-friction graphite pistons are used to remotely drive the needle into the tissue.  During insertion, the surgeon is sensitive to loading changes as loading indicates different phases of needle insertion, particularly skin penetration, and contact with target organ. The linear regression model shown in figure \ref{fig:GraphitePistonMotion}, shows an almost one-to-one motion transmission in the needle insertion axis.  Furthermore, from figure \ref{fig:SystemBlockDiagramv1}, the insertion servomechanism from $z_{12r}$ to $z_{12}$ withe $z_{11r} = 0$, is tuned for a $f=4$ Hz bandwidth.  Similarily, adding the fluid transmission through an $L=15$ ft line, the bandwidth from $z_{12r}$ to $z_{s}$ is about $f=2.25$ Hz. 
\indent As mentioned in previous section, the force transparency of the sits above $85\%$ for network lines up to 30 ft. Furthermore, the transparency suffers only a $30-50\%$ loss for master input forces of at least 0.5 N as shown in figure \ref{fig:GraphitePistForce}.  The glass cylinder's low-friction also results in less than $5\%$ motion transmission loss even for step sizes of $0.2$ mm.  

\subsection{Multi-Mode Control}

\subsubsection{Control Architecture} 
In the current configuration, each axis of motion has an additional master that permits split-axis collaborative assistance or augmentation between human and motorized master units. A modular control architecture that includes motorized servo-controlled fluid actuators for a fault-driven virtual fixture, collaborative penetration, and angular motion compensation. The robotic system is contained in $G_A$ and $G_N$ where the subscripts denote the angle and needle respectively. Fluid network dynamics are characterized by a set of 2-input 4-output plants $G_1,G_2,G_{1N}$.  In the angulation branch, the outputs $y_1,y_2$ are converted to angle domain $\theta_s,\varphi_s$ by forward kinematics. The needle slave displacement, $z_s$ indicates needle tip position. 

\subsubsection{Needle insertion Multi-master collaboration (collaborative penetration)}

\begin{figure}
    \centering
    \includegraphics[width=0.99\linewidth]{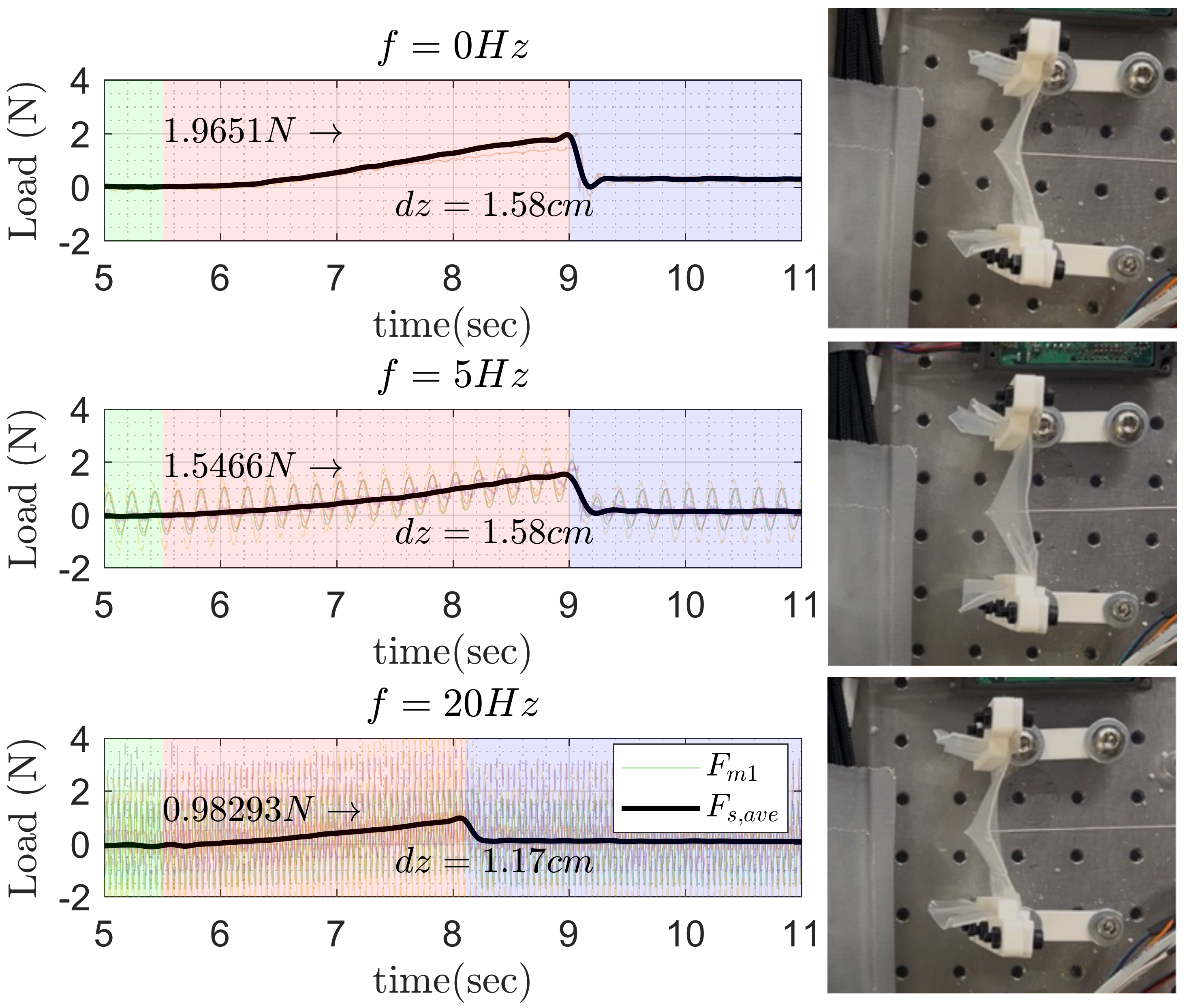}
    \caption{Add-on rapid oscillation at different frequencies. Average peak force and deflection measured by master displacement. (Green) Pre-insertion (Red) Deflection and (Blue) Penetration. }
    \label{fig:JackHammerAssem1}
\end{figure}

The multi-master collaborative feature of the master-slave robot allows for greater functionality and better performance during percutaneous interventions. Stiff tissues can absorb significant force before tearing and thus increase the probability of needle bending or buckling, forcing the surgeon to perform coarse correction maneuvers. The hydrostatic force transmission network allows the second master actuator to superpose a high frequency force input to the manual master's gross motion thus reducing the required average force for tissue penetration.\\
\indent To demonstrate this functionality, the robot is oriented at a fixed angle [\ref{fig:MultiDOF1}].  The state variables, $(x_{11},x_{12}),(x_{12},x_{22}),z_a, \varphi_a$ are fixed leaving only the needle insertion DoF unconstrained.  A manual operator applies a 45 mm/s constant velocity motion input at $z_{11}$ while the motor-operated needle driver $z_{12}$ toggles between a set of uni-modal motion inputs ranging from $0-20Hz$.  The skin phantom is a 4 mm layer of silicon rubber gel that sits taught and  perpendicular to the needle advancement.   The  unassisted penetration results in a 1.97 N peak load prior to needle puncture [\ref{fig:JackHammerAssem1}].  The assistive, digital master's oscillating pressure input, the average peak load and the tissue deflection is reduced. This higher frequency mode serves to partially tear the tissue and thus result in less needle deflection during tissue puncture.  Needle deflection, average reaction force, and average internal fluid pressure decrease $25-50\%$ with the $f=20$ Hz superposed mode.  

\subsubsection{Needle Angulation Multi-master collaboration: Pressure Restriction (Virtual Fixture)}
\begin{figure}
\centering
{\includegraphics[width=0.495\textwidth]{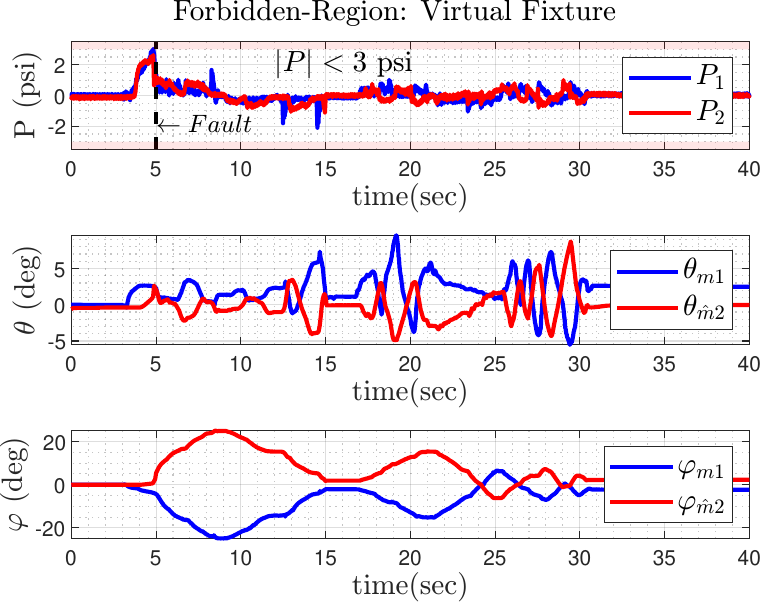} }
\caption{Internal pressure in fluid line rises above $3$ psi and triggers a fault-induced pressure regulation protocol. Resulting motion waveform of digital master $(\theta_{\hat{m}2},\varphi_{\hat{m}2})$ is inverse of human master $(\theta_{m1},\varphi_{m1})$.}\label{fig:VirtualFix}
\end{figure}

Collaboration is necessary as part of the manual teleoperation fault management protocol.  During needle angulation, the interaction between the needle and the needle path environment (tissue, organs, bones) may trigger a large reaction force indicating an undesirable contact force (i.e. $|P|>\sigma$, $\sigma$=maximum allowable pressure).  Inline pressure sensors in the transmission line are scheduled to detect this interaction and throw a fault response.
Under an active fault all motorized servo-mechanisms switch from a position to a pressure servo loop where the latter is configured to regulate pressure to 0 gauge. In this scenario, the motorized master actuators ``absorb" the propagated pressure waves unknowingly generated by the human master end and ``protect" the slave end from exerting force onto the tissue.  Interestingly, when transforming the compensatory motion measured on the digital master side, they result in the inverse waveform of the human-master's input as shown in figure \ref{fig:VirtualFix}. 

\subsubsection{Needle Angulation Multi-master collaboration: Angulation feedback compensation} 
\begin{figure}[h!]
    \centering
    \includegraphics[width=0.95\linewidth]{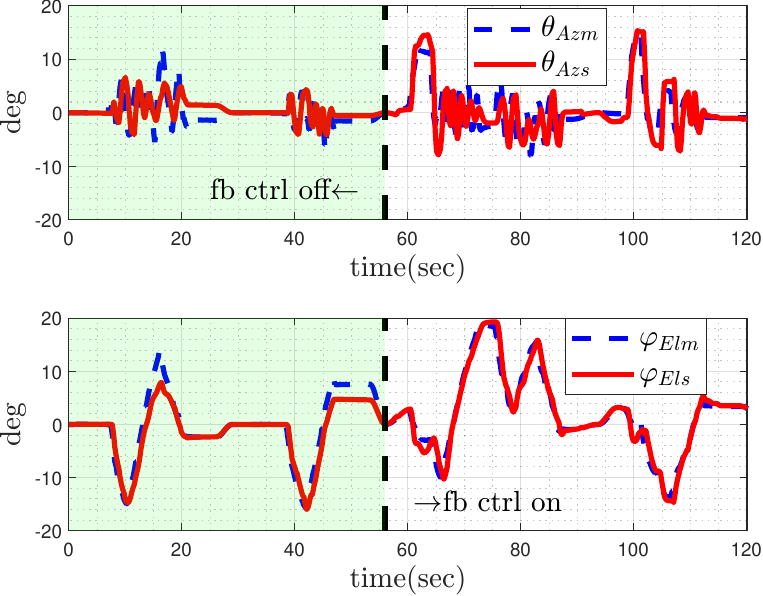}
    \caption{Multi-master collaborative motion tracking performance. Motorized master ($m_2$) leverages real-time slave angular feedback to compensate for angulation motion.}
    \label{fig:MS_AzEl_Fbctrl}
\end{figure}

As described in figure \ref{fig:SystemBlockDiagramv1}, an additional feature of motorized robot collaboration is to perform dynamic needle angle compensation using feedback control.  In this paper, we use an IMU to emulate needle orientation feedback from the MRI. Although many MR image processing modalities are capable of 0.5-10 Hz sample rate image feedback rate, in the bench-top system, we set a 100 Hz angle read-back.  A digital reference command is acquired via the digital joystick seen in figure \ref{fig:MultiDOF1}. Angle error dynamics processed by a SISO linear controller in the angle domain before undergoing a linear transformation, $T^{-1}\approx IK$, to actuator displacement domain. The displacement commands are transmitted to the fluid actuator servo's that drive the master fluid actuators.  In figure \ref{fig:MS_AzEl_Fbctrl}, human master driven response is recorded before the feedback control is turned on at $t=57$ sec.  Improvement in error statistics is described in table \ref{tab:NeedleFbCtrl}. The azimuthal angle $\theta$ is sensitive to small differences in actuator displacements, $|y_1-y_2|$.

\begin{table}[]
    \centering
    \begin{tabular}{llll}
            \toprule
            Axis & rms(deg) & max(deg)& mean(deg)\\
        \midrule
         $\theta$ OL & 3.81 & 14.0 & -1.03 \\ 
         $\theta$ CL & 2.95 & 11.3 & 0.094 \\ 
         $\varphi$ OL & 2.19 & 6.43 & 0.037 \\ 
         $\varphi$ CL & 1.17 & 4.96 & -0.015
    \end{tabular}
    \caption{Improved motion tracking error statistics with multi-master needle angulation compensation.}
    \label{tab:NeedleFbCtrl}
\end{table}

\begin{figure}[h!]
    \centering
    \includegraphics[width=1\linewidth]{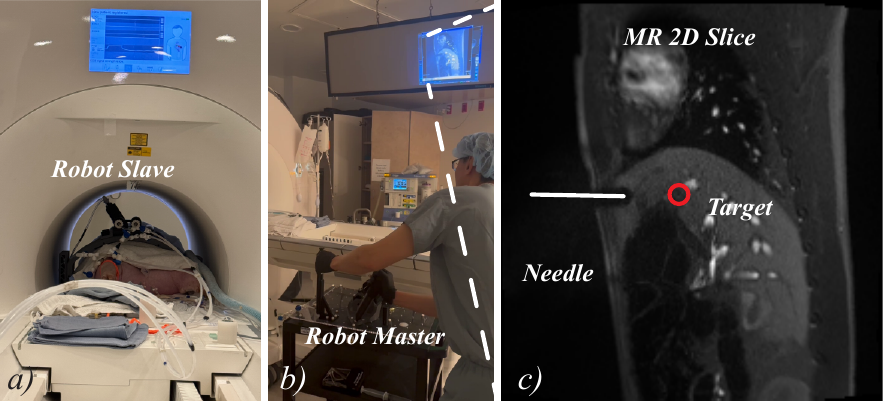}
    \caption{(a) Patient is placed in supine orientation inside a MR bore with MR coil and robot arch over the abdomen. (b) Surgeon remotely manipulates needle angle and insertion depth from bedside. (c) On an overhead projector, 2D sagittal image feedback displays needle and target signature.}
    \label{fig:WorkflowInsert1}
\end{figure}

\section{Human-In-The-Loop Needle Manipulation} \label{sec:HILNM}
The operator proceeds with a 3D MR high resolution preoperative scan to determine the needle entry site. The surgeon investigates the 3-D map to determine an optimal path considering adjacent organs, skin layer density, and local ribs.  After the optimal needle-tip path is decided, the surgeon inserts the needle about 2-2.5 cm deep through the skin layer at a fixed angle and  guides the needle to the target.  Surgeon manipulates the manual master manipulator to adjust the angle of the needle as it progressively advances toward the target. Two-dimensional image feedback displays the in vivo pig torso in the sagittal plane with the needle and target MR signature to guide the surgeon's needle advancements.

As demonstrated in figure \ref{fig:PigExpFinal} and \ref{fig:WorkflowInsert1}, using online image feedback, the surgeon remotely re-positions the needle tip to align the needle axis with the target. Given that the needle insertion module is sensitive to force along the insertion axis, the surgeon leverages the force feedback to infer proximity to the target tissue or confirm contact with neighboring tissue. 

\section{Conclusion}
The table-mounted robot presented here delivers a space-efficient MRI needle angulation and insertion module that overcomes the challenges faced with traditional MRI-conditional robots.  The robot eliminates the need for bedside optical encoders, piezoelectric motors, and ultrasonic motors. Furthermore, it's teleoperation feature also eliminates the need for electric motors that demand additional power electronics to remotely actuate the table-mounted device.  Secondly, traditional actively controlled RCM mechanisms require precise control strategies to maintain an accurate RCM.  Most passive RCM mechanisms are body-mounted and suffer breathing motion perturbations.  The use of a table-mounted fixed center-of-motion ensures that during intra-operative periods, patient body movement does not alter the previous needle state.  The use of high impedance fluid actuators also prevent unexpected forces to affect the needle position and orientation. 

This article presents a teleoperated needle insertion robot manipulator for liver biopsy.  The robotic manipulator's contributions can be summarized as follows.
\begin{enumerate}
    \item The robot is MR safe with minimal effect on MR image noise and distortion. 
    \item The robot mechanism is designed to fit in 60 cm diameter MR bores for in vivo pig experiments and can be feasibly scaled to 70 cm in future iterations. 
    \item The robot is equipped with a hydrostatic actuation network that allows flexible re-configuration while preserving high force transparency and motion transmission.  
    \item The robotic system offers a variety of hybrid and/or collaborative control modalities to assist or augment its functional capabilities such as robot-assisted remote needle insertion and penetration, fault-triggered virtual fixtures, and remote angle feedback motion compensation.
\end{enumerate}

Ongoing and future work includes investigation of MRI-guided closed-loop needle placement using a similar architecture developed in this paper.  For prolonged needle interventions beyond breath-hold time limits, needle orientation and position feedback can be used to compensate for incurred motion errors. The spatio-temporal needle and tissue mechanical interaction motivate integrating adaptive feedback motion compensation strategies for improved performance. Lastly, in addition to the primary surgical use of the needle and leveraging the force-fidelity of the low-friction glass actuator, a remote tissue impedance estimation modality can offer yet another clinically, interpretable information stream for the surgeon.  

{\appendices
\section*{Appendix: Forward Kinematics}
 The forward kinematics of the serial-parallel hybrid master-slave manipulator are treated as cascaded mechanisms with each mechanism's configuration variables determined by the previous link-chain kinematics.
 \subsection*{Robot Forward Kinematics: 4-bar linkage}
 The modified-DH parameters are written for the constrained 4-bar linkage in tables the left and right sub-tables denoted in table \ref{tab:FullDHTable}. The forward kinematics for the multi-bar linkage can be solved by equating frames \textcolor{red}{5} from the \textit{arm 1} and \textit{arm 2}.

\begin{equation}
    ^{i-1}_i\boldsymbol{T} = \boldsymbol{R}_x(\alpha_{i-1})\boldsymbol{D}_{x}(a_{i-1})\boldsymbol{R}_z(\theta_i)\boldsymbol{D}_z(d_i)
\end{equation}
\begin{align}
^{0,r}_{1,r}T{^{1,r}_{2,r}}T{^{2,r}_{3,r}}T{^{3,r}_{4,r}}T{^{4,r}_{5,r}}T = ^{0,l}_{1,l}T{^{1,l}_{2,l}}T{^{2,l}_{3,l}}T{^{3,l}_{4,l}}T{^{4,l}_{5,l}}T
\end{align}
The 4-bar linkage constraint can be separated by rotation and translation.  In the rotational equation, the rotation matrices $\in SO(3)$ show the net angle difference of the linkage closed loop. 

\begin{align}
    R_z(\theta_1)R_z(\varphi_1) &= R_z^T(\theta_2)R_z^T(\theta_3)  \\
    \implies \varphi_1 &= \pi-(\theta_1+\theta _2 +\theta _3) 
\end{align}
In the translation part of the constraint, the remaining configuration variables are obtained. 
\begin{align}
    \rho_1 &= l_1+0.5d_1c(\varphi_1) + D_1s(\varphi_1) \\
    \rho_2 &= 0.5d_1s(\varphi_1)-D_1c(\varphi_1) \\
    \rho_3 &= \rho_1 -0.5c(\pi-\varphi_1)-D_1s(\pi-\varphi_1)\\ 
    \rho_4 &= \rho_2 +0.5s(\pi-\varphi_1)-D_1c(\pi-\varphi_1) \\
    \gamma &=\rho_3^2 +\rho_4^2 - (q_2-q_1)^2-l_2^2-l_{1-2}^2 
\end{align}
A polar coordinate transformation proceeds in order to solve for the remaining configuration variables. 

\begin{align}
    r_1s\beta &= 2(q_2-q_1)l_2 \\
    r_1c\beta &= 2l_{1-2}l_2 \\ 
    \beta & = \tan2^{-1}((q_2-q_1),l_{1-2})\\
    r_2c\gamma_a&=(q_2-q_1)+l_2c\theta_2 \\ 
    r_2s\gamma_a&=l_{1-2}-l_2s\theta_2\\
    \gamma_a &= \tan2^{-1}((l_{1-2}-l_2s\theta_2),((q_2-q_1)+l_2c\theta_2)) \\
\end{align}

\begin{align}
    \theta_2 &= \beta-\sin^{-1}(\gamma/r_1)\\
    \theta_1 &= \gamma_a-\tan2^{-1}(\rho_4,\rho_3) \\
    \theta_3 &= \pi-\varphi_1-(\theta_1+\theta_2) \\
    \phi_1 &= (\theta_1+\varphi_1)-\pi/2
\end{align} \label{eq::FKLinks}

\subsection*{Robot Forward Kinematics: Needle Angulation Pivot Arm }
The robot kinematics governed by the 4-bar linkage drive the needle insertion module \textcolor{blue}{handle}.  The origin lies near the pivot arm.  There are two serial paths for defining the configuration of the needle module's body frame.  The serial paths can be defined by a series of homogeneous transformations that can be defined by appropriated modified DH parameter table. 
The first two frames in the full DH table denote the translation of the arch on the MR bed ($z_a$) and the rotation of the manipulator mount along the arch ($\varphi_a$). 
From frame $o_0$ the forward kinematics are split into two paths: one for the right actuator and the other one for the left actuator. The transformations meet on frame 5 to and continue down to the pivot point before following the path back towards frame $o_0$.  The pivot arm is a spherical joint and is modeled as an $XYZ$ Euler angle sequence ($\alpha,\beta,\gamma_0$).

\begin{align}
^O_A\boldsymbol{T}^A_B\boldsymbol{T}^B_{C_o}\boldsymbol{T}^{C_o}_C\boldsymbol{T} &=\, ^O_{E_i}\boldsymbol{T} ^{E}_{E_i}\boldsymbol{T}^E_{D}\boldsymbol{T}^{D}_{C_i}\boldsymbol{T}^{C_i}_{C}\boldsymbol{T}
\end{align} \label{eq::FK2}

\begin{align} \label{eq::p1}
    \boldsymbol{p}'&=\,^O_A\boldsymbol{R}_z(\phi_1)^T \, ^O_{E_i}\boldsymbol{R}_y(\phi_4)\, ^{E_i}\boldsymbol{p}^{E/O}-^O_{A}\boldsymbol{R}_z(\phi_1)^T\,^O\boldsymbol{p}^{A/O} \\
    \begin{bmatrix}
        p'_x \\ 
        p'_y \\
        p'_z 
    \end{bmatrix} &= \begin{bmatrix}
        (L_1+L_2)c\phi_2+(p_1+L_3)s\phi_2c\phi_3\\
        -(p_1+L_3)s\phi_3 \\ 
        -(L_1+L_2)s\phi_2+(p_1+L_3)c\phi_2c\phi_3
    \end{bmatrix} \\
    \implies p_1 &= +\sqrt{||\boldsymbol{p}||^2-(L_1+L_2)^2}-L_3
\end{align}
Note that in equation \ref{eq::p1}, $L_3<0$ and $p_1<0$.  Furthermore, from inspection, 
\begin{equation}
    \phi_3= \arcsin(-\frac{p_y'}{p_1+L3})
\end{equation}
\begin{align}
    \rho_0c\gamma &= L_1+L_2 \\ 
    \rho_0s\gamma &= -(p_1+L_3)c\phi_3 \\
    c(\gamma +\phi_2) &= \frac{p_x'}{\rho_0} \\ 
    s(\gamma+\phi_2) &= \frac{p_z'}{\rho_0} \\ 
    \gamma & = \arctan2(-(p_1+L_3)c\phi_3,L_1+L_2) \\
    \phi_2 &= \arctan2 (-p_z',p_x')-\gamma
\end{align}

The final step for solving for the forward kinematics is to retrieve the Euler angles that define the needle pivot orientation on the spherical joint.  

\begin{align}
    ^O_CR_{zyx} &=\begin{bmatrix}
        c\beta c\gamma_0 & -c\beta s\gamma_0 & s\beta \\
        s\alpha s\beta c\gamma_0 + c\alpha s\gamma_0 & -s\alpha s\beta s\gamma_0+c\alpha c\gamma_0 & -s\alpha c\beta \\ 
        -c\gamma_0 c\alpha s\beta+ s\alpha s\gamma_0 & c\alpha s\beta s\gamma_0 + s\alpha c\gamma_0 & c\alpha c\beta
    \end{bmatrix} \\ 
    ^O_CR_{zyx} &= ^O_AR_z (\phi_1)^A_BR_y(\phi_2)^B_CR_x(\phi_3)=
    \begin{bmatrix}
        R_{11} & R_{12} & R_{13} \\
        R_{21} & R_{22} & R_{23} \\
        R_{31} & R_{32} & R_{33}
    \end{bmatrix} 
\end{align}

\begin{align}
    \gamma_0 &= \arctan2 (-R_{12},R_{11}) \\ 
    \alpha  &= \arctan2  (-R_{23},R_{33}) \\ 
    \beta &= \arctan2   (R_{13},R_{33}/c\alpha)
\end{align}

\begin{align}
    \theta & = \arctan2(-R_{12},R_{22}) \\ 
    \varphi & = \arctan2(\sqrt{R_{13}^2 + R_{23}^2},R_{33}) 
\end{align}
\section*{Appendix: Inverse Kinematics}
The inverse kinematics involves solving for the fluid actuator displacements, $q_1$ and $q_2$ as a function of the desired needle angle, $\alpha,\beta,\gamma_0$ or in terms of the azimuthal ($\theta$) and elevation ($\varphi$).  In needle biopsy applications $\gamma_0$ is unobservable due to the symmetry of the needle under MR imaging.  Using equation, \ref{eq::FK2}, we can obtain the objective function, $f$. Using equation, \ref{eq::FKLinks} to solve for the unknown, $\theta_1$.  The objective function is constructed as the squared difference of the $p_y$ coordinate associated with frame $A$.  

\begin{align}
    f(\gamma_0) =& ||\boldsymbol{e}_2^T\, (^O\boldsymbol{p}^{A/O,top}-^O\boldsymbol{p}^{A/O,bot})||^2 \\
    \boldsymbol{e}_2^T\,^O\boldsymbol{p}^{A/O,top} &= \frac{l_{1-2}}{2}+ \boldsymbol{e}_2^T\,\boldsymbol{R}_z(\theta_1)\boldsymbol{R}_z(\varphi_1)\begin{bmatrix}
        d_1/2 \\ -D_1
    \end{bmatrix}\\
    \boldsymbol{e}_2^T\,^O\boldsymbol{p}^{A/O,bot} &=-R'_{21}(L_1+L_2)-R_{23}(p_1(\gamma_0)+L_3) \\
    \theta_1 &= \phi_1+\pi/2-\varphi_1
\end{align}

Wolfe conditions
\begin{align}
f(x_k+\alpha_kp_k) \leq f(x_k)+c_1\alpha_k\nabla f_k^T p_k, \\
\nabla f(x_k+\alpha_kp_k)^Tp_k \geq c_2 \nabla f_k^Tp_k, 
\end{align} \label{eq::WolfeCond}

The process of solving for $\gamma_0$ requires solving the nonlinear inverse kinematics for a prescribed needle angle given by $\alpha,\beta$ angles.  The problem can be written as follows 

\begin{equation}
    \min_{\gamma_0} f(\gamma_0)
\end{equation}

We use a BFGS Newton-Step algorithm to solve the nonlinear equation for $\gamma_0$ \cite{Nocedal2006-wj,FletcherOpti}.   The inverse kinematics model is approximated as a quadratic model with matched gradients at each step.

\begin{equation}
    f(\gamma_{0,k+1})\approx f(\gamma_{0,k})+\nabla f(\gamma_{0,k})^T\Delta\gamma_{0,k}+\frac{1}{2}\Delta\gamma_{0,k}^TB_{k+1}\Delta\gamma_{0,k}
\end{equation}

Where the Hessian is approximated as follows 

\begin{equation}
    B_{k+1}\alpha_k\Delta f(\gamma_{0,k})\approx\nabla f(\gamma_{0,k+1}) - \nabla f(\gamma_{0,k}) 
\end{equation} \label{eq::secantEq}

\begin{algorithm} 
\caption{Quasi-Newton $\gamma_0$ solving approach}\label{alg:cap}
    \begin{algorithmic}[1]
    \REQUIRE Starting point $x_0$ and convergence tolerance $\epsilon > 0$, inversion Hessian initial approximation $H_0$, and a step length $\alpha_k$ that satisfies the Wolfe conditions
    \STATE $k \gets 0$; 
        \STATE $\gamma_{0k}=0$ 
        \STATE Initialize $\nabla f_0$
        \STATE $H_0 \gets \frac{y_k^Ts_k}{y_k^Ty_k}I$
    \WHILE{$||\nabla f_k||> \epsilon$}
            \STATE $p_k = -H_k\nabla f_k$
            \COMMENT{Update search direction, $p_k$}
            \STATE $\gamma_{0,k+1}=\gamma_{0,k}+\alpha_k p_k$  \COMMENT{Update $\gamma_{0,k}$ in the search direction. }
            \STATE $s_k=x_{k+1}-x_{k},$ and $y_k=\nabla f_{k+1}-\nabla f_k $; 
            \STATE $\rho_k=\frac{1}{y_k^Ts_k}$
            \STATE $H_{k+1}=(I-\rho_ks_ky_k^T)H_k(I-\rho_ky_ks_k^T)+\rho_ks_ks_k^T$
            \STATE $k \gets k+1$
    \ENDWHILE
    \end{algorithmic}
\end{algorithm}

After the minima is reached from the iterative process, the remaining unknown configuration variables are obtained. Using the rientation block matrix associated \ref{eq::FK2} the internal configuration variables $\phi_1,\phi_2,\phi_3$ are obtained. 

\begin{equation}
    ^O_C\boldsymbol{R}= ^E_D \boldsymbol{R}_x(\alpha)^D_{C_i}\boldsymbol{R}_y(\beta)^{C_i}_C\boldsymbol{R}(\gamma_0)
\end{equation}
\begin{align}
    \phi_1 &= \arctan2(^O_C\boldsymbol{R}_{21},^O_C\boldsymbol{R}_{11}) \\
    \phi_3 &= \arctan2(^O_C\boldsymbol{R}_{32},^O_C\boldsymbol{R}_{33}) \\ 
    \phi_2 &= \arctan2(-^O_C\boldsymbol{R}_{31},^O_C\boldsymbol{R}_{32}/\sin(\phi_3))
\end{align}

\begin{align}
        p_x &= -(L_1+L_2)c(\phi_1)c(\phi_2)-s(\beta)(p_1+L_3)+c(\phi_4)L_4\\
        p_y &= -(L_1+L_2)s(\phi_1)c(\phi_2)+s(\alpha)c(\beta)(p_1+L_3) 
\end{align}

After solving for the position of frame 5, we can then proceed to solve for the fluid actuator displacements, $q_1,q_2$. 

\begin{align}
    \zeta_1 &=\pi-\varphi_1-\theta_1\\
    \zeta_2 &= l_2^{-1}(l_{1-2}/2-p_y-d_1/2s(\zeta_1)+D_1c(\zeta_1)) \\
    \theta_2 &= \arcsin(\zeta_2) \\
    \theta_3 &= \zeta_1-\theta_2 \\ 
    q_1 &= p_x +L_x -l_1c(\theta_1)-d_1/2c(\theta_1+\varphi_1)- D_1s(\theta_1+\varphi_1) \\
    q_2 & = p_x + L_x -l_2 c(\theta_2)-d_1/2(c(\theta_2)c(\theta_3)-s(\theta_2)s(\theta_3)) - \\ & \,\,D_1(c(\theta_2)s(\theta_3)+s(\theta_2)c(\theta_3))
\end{align}

}

\bibliographystyle{IEEEtran}
\bibliography{RobotPaperBib}
\end{document}